\documentclass[conference]{IEEEtran}

\IEEEoverridecommandlockouts

\usepackage{cite}
\usepackage{amsmath,amssymb,amsfonts}
\usepackage{algorithmic}
\usepackage{graphicx}
\usepackage{textcomp}
\usepackage[dvipsnames]{xcolor}

\usepackage{macros}

\begin{document}

\title{Adversarial Reinforcement Learning for Detecting False Data Injection Attacks in Vehicular Routing}

\author{\IEEEauthorblockN{Taha Eghtesad}
\IEEEauthorblockA{\textit{Informatics and Intelligent Systems} \\
\textit{Pennsylvania State University}\\
University Park, USA \\
tahaeghtesad@psu.edu}
\and
\IEEEauthorblockN{Yevgeniy Vorobeychik}
\IEEEauthorblockA{\textit{Computer Science \& Engineering} \\
\textit{Washington University in St. Louis}\\
St. Louis, USA \\
yvorobeychik@wustl.edu}
\and
\IEEEauthorblockN{Aron Laszka}
\IEEEauthorblockA{\textit{Informatics and Intelligent Systems} \\
\textit{Pennsylvania State University}\\
University Park, USA \\
laszka@psu.edu}
}

\IEEEaftertitletext{\centering Published in the Proceedings of the 17th ACM/IEEE International Conference on Cyber-Physical Systems (ICCPS 2026).}

\maketitle

\begin{abstract}
In modern transportation networks, adversaries can manipulate routing algorithms using false data injection attacks, such as simulating heavy traffic with multiple devices running crowdsourced navigation applications, to mislead vehicles toward suboptimal routes and increase congestion. To address these threats, we formulate a strategically zero-sum game between an attacker, who injects such perturbations, and a defender, who detects anomalies based on the observed travel times of network edges. We propose a computational method based on multi-agent reinforcement learning to compute a Nash equilibrium of this game, providing an optimal detection strategy, which ensures that total travel time remains within a worst-case bound, even in the presence of an attack. We present an extensive experimental evaluation that demonstrates the robustness and practical benefits of our approach, providing a powerful framework to improve the resilience of transportation networks against false data injection. In particular, we show that our approach yields approximate equilibrium policies and significantly outperforms baselines for both the attacker and the defender. 
\end{abstract}

\maketitle


\section{Introduction}

The rise of crowdsourced navigation applications, such as Google Maps%
, Waze%
, and Apple Maps%
, has revolutionized urban mobility by offering route suggestions for drivers based on real-time traffic conditions. However, these platforms are increasingly vulnerable to a growing threat vector: false data injection (FDI) attacks~\cite{schoon2020google,eryonucu2022sybil}. These attacks %
aim to deliberately manipulate crowdsourced data to mislead routing algorithms and disrupt transportation networks.

FDI attacks exploit the decentralized and participatory nature of crowdsourced navigation systems. 
For example, researchers demonstrated such an attack by placing mobile devices running Google Maps in a cart and slowly dragging it along a street~\cite{schoon2020google}. The system misinterpreted the data as indicating severe congestion
and rerouted vehicles away from the area. This illustrates how easily attackers can deceive these systems using relatively simple techniques.
In addition to manipulating crowdsourced data, attackers may also target physical traffic sensors, exploiting the poor state of cybersecurity in many transportation deployments.
For example, tampering with vehicle-counting sensors or signal-timing systems can distort traffic data and amplify the impact of FDI attacks%
\hbox{\cite{feng2018vulnerability,jamalzadeh2022protecting}}.

FDI attacks pose a serious threat to transportation networks. By injecting false data, %
attackers can trick navigation systems into rerouting countless vehicles, creating widespread gridlock. The most critical impact of this deliberate congestion is the potential to delay emergency responders, where even minutes lost can have life-threatening outcomes. Beyond this critical impact, cascading effects include longer commute times, increased fuel consumption and vehicle emissions, and disruptions to public transit and logistics, inflicting a significant economic and environmental toll. %

One approach to mitigating FDI attacks is to promptly detect them based on anomalous traffic observations, complementing traditional cybersecurity measures, such as access control. Traffic observations can include measurements of physical variables, such as the congestion levels or estimated traffic speeds of specific road segments. By comparing these traffic observations with data recorded during normal operations, it is possible to identify anomalies that indicate an attack. Classical anomaly detection methods, such as applications of statistical and machine learning models, could play a crucial role in identifying unusual traffic patterns. %

However, attackers may attempt to carry out stealthy attacks that maximize disruption while minimizing the likelihood of detection. Such stealthy attacks aim to carefully manipulate data to remain within the range of normal variation, thereby evading classical anomaly detection methods. In addition, attackers might adapt their tactics if they are aware that the detector was trained on previously executed attacks, further complicating detection efforts, which need to anticipate such adaptive attacks. The literature has not yet explored strategies to counter such stealthy and adaptive attacks, leaving a critical gap in current defenses.

These challenges underscore the need for a robust detection mechanism against FDI attacks on transportation, capable of thwarting the attackers' efforts to avoid detection.
Developing these capabilities is essential for enhancing the resilience of crowdsourced navigation systems against evolving threats.

Recent research has focused on identifying and simulating effective attack strategies. Eghtesad \emph{et al.}~\shortcite{eghtesad2024multi} introduced a scalable reinforcement learning algorithm that generates attack strategies capable of maximizing network disruption. Yang \emph{et al.}~\shortcite{yang2024strategic} and Yu \emph{et al.}~\shortcite{yu2024sensing} explored the impact of random attacks, where false demands are distributed across network edges using uniform or Gaussian distributions, and proposed defenses such as trusted, unperturbed sensors to mitigate these threats. However, identifying vulnerabilities is only the first step. The critical open challenge, which we address in this work, is the design of a robust detection mechanism that can withstand strategic, adaptive attackers.
We propose a robust detection model that identifies attacks based on the observed travel times of network edges, anticipating a strategic attacker who responds by launching a worst-case stealthy attack.

\paragraph{Contributions}
Our contributions are threefold: (1) we formulate a strategic zero-sum game between an attacker employing FDI and a defender detecting such attacks, (2) we demonstrate that solving for the Nash equilibrium of this game provides the optimal detection strategy, ensuring resilience against adaptive adversarial strategies, and (3) we leverage a policy-space response algorithm to compute equilibrium strategies efficiently, providing an optimal detection mechanism.
This approach ensures minimal disruption to the transportation network under a worst-case attack, thereby enhancing its robustness against FDI manipulations.

\paragraph{Organization}
Section~\ref{sec:related} reviews related work on attacks against transportation networks and on reinforcement learning for cyber defense. Section~\ref{sec:prelim} covers the theoretical background of our approach. Section~\ref{sec:system} introduces our model of transportation networks and false data injection attacks. Section~\ref{sec:model} presents our game-theoretic detection model and our computational approach. Section~\ref{sec:result} demonstrates the effectiveness of our approach through experiments. Finally, Section~\ref{sec:conclusion} provides concluding remarks. %

\section{Related Work}
\label{sec:related}

The vulnerability of transportation networks to various attacks has been extensively studied, and the application of reinforcement learning in attack detection has been explored in other domains. However, the detection of attacks on crowdsourced navigation applications remains underexplored. \change{Our work addresses this gap by leveraging reinforcement learning to develop an effective detection framework.}{}

\paragraph{Attacks Against Transportation Networks}
\label{sec:related_attacks}

Malicious actors can exploit vulnerabilities in transportation networks to manipulate drivers’ route choices using methods such as sending malicious SMS messages \cite{waniek2021traffic}, physically altering road signs \cite{eykholt2018robust}, tampering with traffic signals \cite{laszka2016vulnerability,feng2018vulnerability}, or injecting false data into crowdsourced navigation applications \cite{eryonucu2022sybil,schoon2020google,eghtesad2024multi,yu2024sensing,yang2024strategic}.
Prior work has explored the vulnerabilities of navigation applications to adversarial manipulations and their impacts on traffic congestion \cite{waniek2021traffic,raponi2022road,yang2024strategic,yu2024sensing}. Further, Eghtesad \emph{et al.}~\shortcite{eghtesad2024multi} have developed a reinforcement learning-based framework that determines the worst-case impact of FDI attacks on navigation applications.

\paragraph{Reinforcement Learning for Detection}
\label{sec:related_rl}

Reinforcement learning is a promising tool for the detection and mitigation of adversarial attacks. Several prior works have explored methods for detecting and mitigating various types of  attacks using reinforcement learning. For example, Sargolzaei \emph{et al.}~\shortcite{sargolzaei2020detection} focus on the detection and mitigation of false data injection attacks in distributed control systems. Malialis \emph{et al.}~\shortcite{malialis2015distributed} examine network intrusion detection techniques using reinforcement learning, while Hu \emph{et al.}~\shortcite{hu2019reinforcement} leverage reinforcement learning to address zero-day attacks.
Finally, Tong \emph{et al.}~\shortcite{tong2020finding} investigate strategies for prioritizing relevant alerts in network intrusion detection systems using multi-agent reinforcement learning.

\section{Preliminaries}
\label{sec:prelim}

The strategic conflict between an attacker and a defender can be modeled as a stochastic game involving the two players. In this framework, the attacker selects an attack policy, while the defender selects a detection policy. 
When one player commits to a policy, their opponent’s strategic choice %
transforms into a Markov decision process.
If the policies form an equilibrium, neither player is incentivized to deviate and pursue an alternative policy.

\paragraph{Markov Decision Processes (MDP)}
\label{sec:prelim_rl}

The tuple $\langle S, A, R, \allowbreak T \rangle$ defines an MDP, where $S$ represents the state space, $A$ represents the action space, $R(s^t, a^t) \mapsto r^t \in \mathbb{R}$ is the reward function that determines the reward $r^t$ received for taking action $a^t \in A$ %
in state $s^t \in S$  at time step $t$, and $T(s^t, a^t, s^{t+1}) \mapsto [0, 1]$ specifies the probability that taking action $a^t$ in state $s^t$ will result in a transition to state $s^{t+1} \in S$ at the next time step $t + 1$.

A \textbf{policy} function in an MDP is a plan of action $\pi(s^t) \mapsto a^t$ that an agent (i.e., a player) will follow. The utility value $u(\pi)$ of a policy function $\pi$ can be expressed as the discounted sum of rewards that the agent will receive by following the policy:
\begin{align}
\textstyle
    u(\pi) = \mathbb{E}\left[\sum_{t=0}^\infty \gamma^t \cdot r^{t} ~\middle|~ \pi\right]
    \label{eq:rl}
\end{align}

A solution to an MDP is a policy that maximizes the utility of the agent: $\pi^* = \argmax_\pi u(\pi)$. We let $u^* = u(\pi^*)$ denote the utility of this optimal policy. A reinforcement learning (RL) algorithm can be used to computationally find such a policy.

\paragraph{Stochastic Games}
\label{sec:prelim_game_theory}

We can define a stochastic two-player game between the attacker and the defender. This game can be expressed as a zero-sum game $G = (\Pi_p, \Pi_{\bar{p}}, u)$, where players choose an MDP \textit{policy} as their \textbf{pure strategy} $\pi_p \in \Pi_p$ from their strategy set $\Pi_p$, i.e., set of all policies available to them, as their strategy to play. The utility of the game when player $p$ follows $\pi_p$ and player $\bar{p}$ follows $\pi_{\bar{p}}$ is denoted $u(\pi_p, \pi_{\bar{p}}) \mapsto \mathbb{R}$. Player $p$ tries to maximize, while their opponent $\bar{p}$ tries to minimize utility by choosing a policy. 

We can also define a \textbf{mixed strategy} $\sigma_p$ as a probability distribution over the player's pure strategy set $\Pi_p$, where $\sigma_p(\pi_p)$ is the probability of the player choosing $\pi_p\in \Pi_p$. The expected utility %
given mixed strategies $\sigma_p$ and $\sigma_{\bar{p}}$ is:  
\begin{align}
\textstyle
u(\sigma_p, \sigma_{\bar{p}}) = \sum_{\pi_p \in \Pi_p} \sum_{\pi_{\bar{p}} \in \Pi_{\bar{p}}} \sigma_p(\pi_p)\sigma_{\bar{p}}(\pi_{\bar{p}})u(\pi_p, \pi_{\bar{p}}).
\end{align}

A pair of mixed strategies $\sigma_p^*, \sigma_{\bar{p}}^*$ for a zero-sum game form a \textbf{mixed strategy Nash equilibrium} (MSNE) \textit{iff}:
\begin{align}
u(\sigma_p, \sigma_{\bar{p}}^*) &\leq u(\sigma_p^*, \sigma_{\bar{p}}^*) \leq u(\sigma_p^*, \sigma_{\bar{p}}) \quad \forall \sigma_p, \sigma_{\bar{p}}
\label{eq:equilibrium}
\end{align}

MSNE ensures that neither player $p$ nor their opponent $\bar{p}$ can improve their expected utility by unilaterally deviating from $\sigma_p^*$ or $\sigma_{\bar{p}}^*$, respectively. \textit{Hence, each player is disincentivized from deviating from their MSNE strategy.}

Given the strategies and utility values, we can compute the MSNE of such zero-sum stochastic games using linear programming~\cite{shoham2008multiagent}.

\paragraph{Double Oracle (DO)}
\label{sec:prelim_do}

The MSNE strategies in zero-sum games with large strategy sets, where the enumeration of pure strategies is infeasible, can be calculated using the DO algorithm~\cite{mcmahan2003planning}.
The algorithm begins by initializing a small subset of the strategy set for each player, denoted $\Pi_p^0 \subset \Pi_p$ and $\Pi_{\bar{p}}^0 \subset \Pi_{\bar{p}}$. At each iteration~$i$, the algorithm computes an MSNE, denoted $\sigma_p^{*,i}$ and $\sigma_{\bar{p}}^{*,i}$, of the subgame $G^i$ that is defined by the current strategy sets $\Pi_p^i$ and $\Pi_{\bar{p}}^i$.

Once an MSNE is computed, each player calculates their \textbf{best response} (BR) pure strategy $\pi_p^{i+1} = \argmax_{\pi_p \in \Pi_p} u(\pi_p, \sigma^{*, i}_{\bar{p}})$ or $\pi_{\bar{p}}^{i+1} = \argmin_{\pi_{\bar{p}} \in \Pi_{\bar{p}}} u(\sigma^{*, i}_{p}, \pi_{\bar{p}})$, respectively, which is the pure strategy that maximizes or minimizes their utility given the mixed strategy $\sigma^{*, i}_{\bar{p}}$ or $\sigma^{*, i}_{p}$ of their opponent. These best response strategies from the players' strategy sets ($\Pi_p$ and $\Pi_{\bar{p}}$) are then added to the player's current strategy sets ($\Pi_p^{i}$ and $\Pi_{\bar{p}}^{i}$):
\begin{align}
\Pi_p^{i+1} \gets \Pi_p^{i} \cup \{\pi_p^{i+1}\} \quad \text{ and } \quad \Pi_{\bar{p}}^{i+1} \gets \Pi_{\bar{p}}^{i} \cup \{\pi_{\bar{p}}^{i+1}\}
\end{align}

The process is repeated iteratively, updating the strategy sets in each iteration. The algorithm continues until the MSNE utility of the subgames $G^i$ converges to the MSNE of the game $G$~\cite{mcmahan2003planning}.

\paragraph{Policy Space Response Oracles (PSRO)}
\label{sec:psro}

In games with enormous policy spaces (i.e., strategy sets), such as video games, it is infeasible to enumerate all the policies when searching for a best response. 
Further, given the opponent is committed to a policy, it is still infeasible to solve the resulting MDP for the opponent using RL.

The PSRO algorithm \cite{lanctot2017unified} extends DO by using Deep Reinforcement Learning (DRL) as an \textit{approximate best response oracle}, which finds an approximate---due to the approximation nature of deep neural networks and reinforcement learning---best response to the opponent's MSNE strategy within the player's pure strategy set: \hbox{$\pi_p^{i+1} \!\approx\! \argmax_{\pi_p \in \Pi_p} u(\pi_p, \sigma^{*, i}_{\bar{p}})$} and $\pi_{\bar{p}}^{i+1} \!\approx\! \argmin_{\pi_{\bar{p}} \in \Pi_{\bar{p}}} u(\sigma^{*, i}_{p}, \pi_{\bar{p}})$. In this framework, each player (i.e., both the attacker and the defender) selects the parameters of a DRL policy as their pure strategy within the game. The strategy set that is defined by all possible DRL policies %
can also be referred to as the \textit{policy space}, which serves as the basis for the term \textit{policy space response oracles}.

\section{System and Threat Model}
\label{sec:system}

\label{sec:system_env}

The road network is represented as a directed graph $G = (V, E)$, where nodes denote intersections and edges denote road segments. Each edge $e = (u, v) \in E$, where $u, v \in V$, is characterized by a \textbf{free-flow travel time} $f_e$ (time to traverse the road segment without traffic), a \textbf{capacity} $c_e$ (maximum traffic flow rate of the segment), and two \textbf{congestion parameters}, $b_e$ and $p_e$, which control travel time under congestion. %
When $n_e$ vehicles are traveling on edge $e$, the travel time $W_e(n_e)$ of the edge is calculated using the Bureau of Public Roads (BPR) function~\cite{transportationnetworks,states1964traffic}:
\begin{align}
    W_e(n_e) = f_e \cdot \left(1 + b_e \left(n_e\mathbin{/}c_e\right)^{p_e}\right).
\end{align}

The travel demand within the network is defined by a set of trips~$R$. Each trip $r \in R$ is a tuple $r = \langle o_r, d_r, s_r \rangle$, representing a demand of $s_r$ vehicles traveling from an origin node $o_r \in V$ to a destination node $d_r \in V$. Through simulation, we can calculate the number of time steps $T_r$ for vehicle trip $r$ to reach its destination.

Consistent with recent prior work \cite{eghtesad2024multi}, we employ a dynamic, agentic step-wise simulation of traffic, where vehicles make realistic routing decisions sequentially over time, a departure from classical static network-flow models.

\subsection{States and Transitions}
\label{sec:system_state}

The system evolves in discrete time steps $t$. The \textbf{state} is the set of all trip locations $\{l_r^t \in V \cup (E \times \mathbb{N})\}$, where a location can be a node $v \in V$ or a position on an edge $\langle e, \tau \rangle$ with $\tau \in \mathbb{N}$ time remaining.

When a vehicle trip $r$ is at a node ($l_r^t = u$), it dynamically routes based on a \change{\textbf{softmin}}{stochastic} policy.
This stochastic policy models limited rationality for a more realistic simulation, representing vehicles that may not completely rely on the navigation application for route planning. In this model, each vehicle calculates the cost-to-go via each adjacent node $v \in N(u)$ as $C_v = w_{(u,v)}^t + d(v, d_r)$, where $d(v, d_r)$ is the shortest path distance based on travel times $w_e^t$. The probability of choosing edge $(u,v)$ follows a Boltzmann distribution:
\begin{align}
\label{eq:path_cost}
P(l_r^{t+1} = \{(u,v), \lfloor w_{(u, v)}^t \rceil\}) = \frac{e^{-\theta C_v}}{\sum_{v' \in N(u)} e^{-\theta C_{v'}}},
\end{align}
where the parameter $\theta \ge 0$ controls rationality; as $\theta \to \infty$, the policy approaches the deterministic choice of shortest path.

Then, the locations are updated for $t+1$. A vehicle trip at a node $u$ moves onto its chosen edge $(u, v)$. A vehicle trip on an edge $\langle e, \tau \rangle$ updates its state: if $\tau>1$, its new state is $l_r^{t+1}=\langle e, \tau-1 \rangle$; if $\tau=1$, it arrives at the edge's destination node, $l_r^{t+1} = v$.

\subsection{Threat Model}
\label{sec:system_threat}

The attacker perturbs the observed travel times of the edges. Let $\vect{a}^t = \{a_e^t \in \mathbb{R}\}_{\forall_{e \in E}}$ denote the perturbations for each edge $e$. The observed travel time that is used to choose the path (\cref{eq:path_cost}) is:
\begin{align}
    \hat{w}_e^t = w_e^t + a_e^t.
\end{align}

Our threat model diverges from Eghtesad~\emph{et al.}~\cite{eghtesad2024multi} by removing the explicit budget constraint on the attack. We argue that this constraint is superfluous because a rational attacker, seeking to maximize expected impact, is already incentivized to self-limit their attack magnitude to balance effectiveness with the inherent risk of detection. An increase in the total magnitude of the attack increases the probability of the attack being detected and thwarted.
Also note that while this threat model permits independent edge modification, an attacker facing the risk of detection is also incentivized to respect traffic flow inter-dependencies. To remain stealthy, an attacker has to mimic legitimate traffic patterns; a sudden change on an isolated edge would be easily detectable.

During an FDI attack, vehicles use $\hat{w}_e^t$ to compute shortest paths. The attacker, assumed to fully observe the environment including the network topology and the vehicle trip origins, destinations, and locations, seeks to maximize the total vehicle travel time:
\begin{align}
\label{eq:attacker_objective}
\textstyle
    u^*_a = \max_{\left\{\vect{a}^1, \vect{a}^2, \ldots\right\}} ~ \sum_{r \in R} s_r \cdot T_r.
\end{align}

\section{Game Theoretic Threat Detection}
\label{sec:model}

\change{In this section, we}{We} present a model for detecting abnormal traffic patterns by formulating the system as a Partially Observable MDP (POMDP) based on the observed travel times~$\hat{\vect{w}}$. Building on this, we extend the model to capture the interaction between an attacker attempting to manipulate traffic patterns by injecting perturbations into the navigation application and a detector aiming to identify such anomalies. This interaction is formulated as a two-player strategic zero-sum game, capturing the adversarial dynamics. To determine the optimal strategies for both players, we leverage the PSRO algorithm to compute the MSNE of the game. The Nash equilibrium guarantees that any non-equilibrium policy will result in suboptimal outcomes: any other attacker strategy will lead to lower total travel time, while any other defense strategy will lead to increased total travel time.

\subsection{Defense Model}
\label{sec:model_detection}

The defender observes the perturbed travel times $\vect{\hat{w}}^t$ at each time step~$t$, and makes a decision whether to raise an alert or not:
\begin{align}
\label{eq:defender_action}
\textstyle
d^t \in \{0, 1\}.
\end{align}
In the event of no attack ($\vect{a}^t = \vect{0}$), the defender observes the nominal travel times $\vect{w}^t$, but cannot %
distinguish them
from the perturbed times $\vect{\hat{w}}^t$ observed under attack (without anomaly detection).

If the defender detects an attack at time $t$ and raises an alert ($d^t =1$), then all future perturbations are prevented, i.e., $\vect{a}^\tau = 0$ for all remaining steps $\tau \geq t$.
In a real-world deployment, detection would trigger mitigation protocols, such as blocking crowdsourced data from untrusted sources or reverting routing logic to historical averages and data from trusted physical sensors.
Failure to detect an ongoing attack ($d^t =0$) enables the attacker to increase vehicle travel times $T_r$ by continuing to perturb observed travel times $\vect{\hat{w}}^t$ (\Cref{sec:system_threat}). %

The defender's objective is to minimize the total vehicle travel  time, that is, to minimize the attacker's utility.
In addition, false positives (i.e., alerts raised when there is no real attack) incur a fixed cost of $C_f$.
\change{By \textit{detecting an attack correctly at timestep $t$}, the defender prevents further perturbations of the edge travel times $\forall_{\tau \geq t}: \vect{a}^\tau = 0.$}{}
We can formulate the defender's objective to minimize the total travel time while minimizing the cost of false alarms:
\begin{align}
\textstyle
    u_d^* = \min_{\{d^1, d^2, \cdots\}} \sum_{r \in R} s_r \cdot T_r +  C_f \cdot |F|,
    \label{eq:defender_objective}
\end{align}
where $|F|$ is the number of false positive alerts raised by the defender before an attack is correctly detected, and $C_f$ is the cost of false positive alerts (i.e., cost of wasted time and effort). By incorporating a false positive cost, the defender can automatically manage the trade-off between false positive rate and total travel time.
This is crucial since in practice,  anomalies may occur without an adversary (e.g., congestion caused by an accident or large public events). Because the defender’s objective function includes an explicit penalty for false positives, an optimal detector can learn to ignore legitimate congestion patterns that lack the specific intent or temporal signature of an FDI attack.

Unlike static anomaly detectors that learn a fixed baseline of normal behavior, our defender is trained against an adaptive attacker, learning to identify sophisticated and adaptive attack patterns by observing the history of travel times ($\hat{w}^t$), which reflect both the attacker's direct perturbations and their indirect, cascading effects on traffic congestion.

This challenge is amplified because observed traffic deviates from normal patterns in two ways. First, deviations arise directly from the attacker's perturbations, which alter the travel-time data used by vehicles for routing. Second, these manipulated observations cause vehicles to reroute, leading to indirect, cascading effects that produce real, yet anomalous, traffic congestion. The defender must therefore untangle anomalies caused by both the attacker's false data and the real-world consequences of those manipulations.

\subsection{The Detection Game}
\label{sec:model_game}

We model the interaction between the attacker and the defender as a two-player game. The attacker maximizes its utility, which is the total travel time, while the defender aims to minimize it by correctly detecting adversarial perturbations while reducing the false positive alarm rate. This game is not strictly zero-sum due to the defender’s \textit{false alarm penalty} (Eq.~\ref{eq:defender_objective}). However, the defender's false alarm penalty depends only on its own strategy and is independent of the attacker’s actions. Except for this penalty, the players’ goals ($u_a$ vs. $u_d$) are directly opposing: any gain for the attacker results in an equivalent loss for the defender in terms of total travel time. As such, the interaction is strategically equivalent to a zero-sum game. To solve the detection game, we employ PSRO (Sec.~\ref{sec:prelim_do}),
iteratively updating the policy space of both players by identifying the best response of one player to the current MSNE strategy of the other.

\begin{figure}
    \centering
    \includegraphics{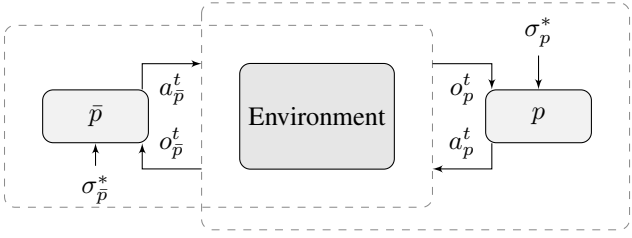}
    \caption{Dividing a two-player game environment into two single-agent reinforcement learnings for the PSRO algorithm: one for player $p$ and one for its opponent $\bar{p}$, where $\bar{p}$ plays with MSNE $\sigma_{\bar{p}}^*$ and $p$ plays with $\sigma_p^*$, respectively.}
    \label{fig:psro}
\end{figure}

\subsection{DRL as Approximate Best-Response Oracle}

In line with PSRO, we use DRL algorithms to find a player's approximate best response to their opponent's current MSNE strategy.

\subsubsection{Attack Oracle}
\label{sec:attack_oracle}

\begin{figure*}[t!]
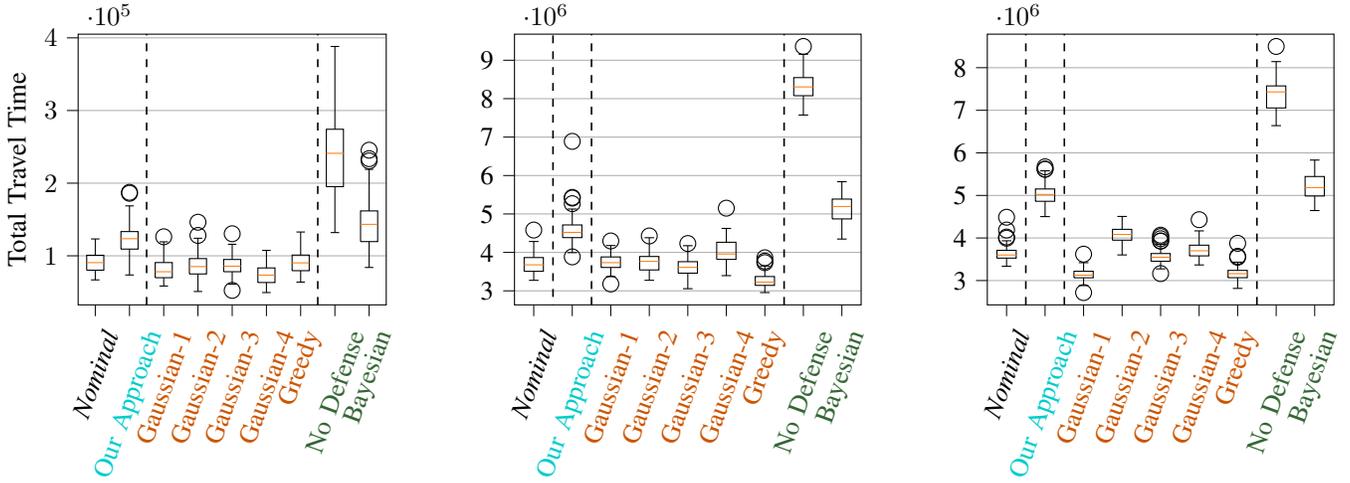

\centering
    \begin{subfigure}[b]{.3\textwidth}
        \centering
        \includegraphics[width=\textwidth,height=6.5cm]{new_plots/3x2/payoff_comparison_all.tikz}
        \caption{3x2 GRE Graph: 6 Nodes, 16 Edges}
    \end{subfigure}
    \hfill
    \begin{subfigure}[b]{.3\textwidth}
        \centering
        \includegraphics[width=\textwidth,height=6.5cm]{new_plots/5x4/payoff_comparison_all.tikz}
        \caption{5x4 GRE Graph: 20 Nodes, 55 Edges}
    \end{subfigure}
    \hfill
    \begin{subfigure}[b]{.3\textwidth}
        \centering
        \includegraphics[width=\textwidth,height=6.5cm]{new_plots/sf/payoff_comparison_all.tikz}
        \caption{Sioux Falls, SD: 24 Nodes, 76 Edges}
    \end{subfigure}
    \hfill
    
    \caption{
    Experimental evaluation and comparison of \textcolor{solution}{our approach}\colorlegendbox{solution} to alternative strategies. %
    First, we compare various baseline \textcolor{attack}{attack strategies}\colorlegendbox{attack} to our \textcolor{solution}{equilibrium solutions};  here, a higher total travel time would indicate a more effective attack (\textbf{higher is better}).
    Second, we compare various \textcolor{defense}{defense strategies}\colorlegendbox{defense} to our \textcolor{solution}{equilibrium solutions}; here, a lower total travel time would indicate a more effective defense (\textbf{lower is better}).
    The results show that \textcolor{solution}{our approach}\colorlegendbox{solution} outperforms all of the alternatives, inducing higher travel times than other \textcolor{attack}{attacks} and securing lower travel times than other \textcolor{defense}{defenses}. The results demonstrate that \textcolor{solution}{our approach}\colorlegendbox{solution} outperforms the alternatives in both roles; crucially, our equilibrium-based defender is robust against these alternative \textcolor{attack}{attacks} without prior training on them, which demonstrates the success of our algorithm.
    }
    \label{fig:comparison}
\end{figure*}

The attacker's DRL oracle finds a policy to perturb the traffic observations, thereby rerouting the vehicles and increasing the total travel time. Any continuous action DRL algorithm, such as the Deep Deterministic Policy Gradient (DDPG)~\cite{lillicrap2015continuous} or Proximal Policy Optimization (PPO)~\cite{schulman2017proximal}, may be applicable. %

The attacker fully observes the traffic environment, including the network's topology and the vehicles' locations and route choices. This information must be engineered into useful machine-learning features that can be used as a state representation for the DRL oracle.
We adopt five edge features from prior work~\cite{eghtesad2024multi}:
the number of vehicles on any node whose unperturbed shortest paths include edge $e$
($s_e^t$), the number of vehicles immediately taking the edge ($\hat{s}_e^t$), the number of vehicles currently on the edge ($m_e^t$), the number of vehicles on any edge whose shortest paths include the edge ($\tilde{s}_e^t$), and the number of vehicles currently on the edge ($n_e^t$). %
Additionally, we incorporate two new edge features based on the network topology: \textbf{edge capacity} ($c_e$) and \textbf{free-flow time} ($f_e$).
Together, these edge features form the attack oracle's observation vector: 
\begin{align}
    \label{eq:attacker_observation}
    \mathbf{o}^t_a = \left\langle\langle s_e^t, \hat{s}_e^t, m_e^t, \tilde{s}_e^t, n_e^t, c_e, f_e \rangle:~ \forall_{e \in E}\right\rangle.
\end{align}

The attacker's action is the observation perturbation applied to each edge $e$ of the network: 
\begin{align}
    \label{eq:attacker_action}
    \vect{a}^t = \langle a^t_e \,|\, a_e^t \geq 0:\forall_{e \in E}\rangle.
\end{align}

\change{As a result of this action, vehicles will be directed on suboptimal routes and will increase the total travel time ($T_r$) of vehicle trip $r$.}{} To translate the attacker's objective (\cref{eq:attacker_objective}) into an equivalent stepwise reward signal, we can reward the adversarial DRL agent by the number of vehicles that are still traveling through the network:\Aron{overloading symbol $r$ (reward $r$ vs. trip $r$) may be confusing}
\begin{align}
    r_a^t = \sum \{s_r \,|\, l_r^t \neq d_r : \forall_{r \in R}\}.
\end{align}

\subsubsection{Defense Oracle}
\label{sec:defense_oracle}

On the defense side, the oracle is simpler in comparison since it outputs a binary decision:
whether to raise an alarm or not (\cref{eq:defender_action}). Consequently, RL algorithms such as Deep Q-Networks (DQN)~\cite{mnih2013human} or PPO are sufficient to approximate the defender’s best response. The reduced complexity of the defender’s decision space ensures that these methods are computationally efficient and effective in finding optimal detection policies.

The reinforcement learning algorithm for the defender %
observes the perturbed edge travel times of the network, same as the vehicles:
\begin{align}
    \label{eq:defender_observation}
    \vect{o}^t_d = \vect{\hat{w}}^t = \left\langle \hat{w}_e^t 
    : \forall_{e \in E}\right\rangle.
\end{align}

The defender operates in a partially observable MDP; to improve the efficiency of DRL, we provide the defender with a history of observations $\vect{h}$, where $\vect{h}_t = \langle \vect{o}^{t-|H|}_d, \cdots, \vect{o}^t_d\rangle$. We set $|H|=5$ in our experiments. The defender's policy outputs a binary action, representing the detection decision at time step $t$:
    $a_d^t \in \{0, 1\}.$
Finally, the defender receives a reward based on the number of vehicles that are still traveling through the network (with a negative sign, so that the number is minimized) and a penalty $-C_f$ for a false alarm since reinforcement learning maximizes rewards:
\begin{align}
    r_d^t &= - \sum\left\{r_s\,|\,l_r^t \neq d_r:\forall_{r\in R}\right\} - p \\
    \nonumber
    p &= \begin{cases}
        C_f \quad &\text{if there is no attack at time step $t$ and $d^t =1$} \\
        0 \quad &\text{if an attack is correctly detected or $d^t = 0$.}
    \end{cases}
\end{align}

\subsection{Solving the Strategic Detection Game}

By iteratively generating approximate best-responses using DRL (see Sec.~\ref{sec:psro}) against the opponents' current MSNE strategies, the DO algorithm refines the strategy sets until convergence.

We employ PPO~\cite{schulman2017proximal} as the DRL algorithm for both the attack and defense best-response oracles. For the attack oracle, PPO optimizes a multivariate Gaussian action distribution with a diagonal covariance matrix to output continuous perturbation values, which are then exponentially scaled to ensure that they are positive. For the defense oracle, PPO optimizes a Bernoulli action distribution for the binary decision of whether to raise an alarm or not. 

DO is guaranteed to converge to an MSNE of the game with exact BR oracles~\cite{mcmahan2003planning}, but not with approximate BR oracles.
We show empirically that our DRL-based approximate BR oracles are general and effective for both the adversary and the defender, so in practice, DO converges to an MSNE in only a few iterations with our oracles. This demonstrates the feasibility and practicality of the proposed approach for real-world traffic monitoring systems.

\section{Experimental Analysis}
\label{sec:result}

Network topologies and vehicle data, such as origins, destinations, and counts, are provided by the Transportation Networks for Research Core Team~\shortcite{transportationnetworks}. We used Sioux Falls, SD as the testbed for our approach. Further, we generated two experimental networks using the Grid model with Random Edges (GRE)~\cite{peng2014random}. GRE stochastically generates a random graph with similar characteristics as a real-world transportation network. For this network, we also generated the edge attributes, e.g., capacity and free flow time, based on the same distribution as Sioux Falls, SD~\cite{transportationnetworks}.
In all cases, vehicle counts for each origin-destination pair are randomized by $\pm 0.05\%$.

The players' initial strategy sets of the PSRO contain only one policy of \textcolor{attack}{No Attack} \colorlegendbox{attack} and \textcolor{defense}{No Defense} \colorlegendbox{defense}, where both players  take zero actions and the environment operates under \textit{\textbf{nominal}} conditions. In each iteration of PSRO, we first train an adversarial DRL and then a detection DRL.

We used the default hyperparameters from Stable Baselines~3~\cite{stable-baselines3}, which are validated by prior work~\cite{andrychowicz2020what}, and left random seeds to their default values in NumPy
and PyTorch,
as our algorithm exhibits low sensitivity to random initialization. Further, to verify the significance of our results, after the Nash equilibrium model is trained, we collect 64 episodic rewards and run permutation statistical tests on them, comparing the means of the distributions. 

\subsection{Baselines}
On these two network models and the Sioux Falls, SD model, we compare our solution approach to attack baselines such as the greedy attack~\cite{eghtesad2024multi} and the Gaussian \cite{yu2024sensing} attack. We also compare our defense approach with state-of-the-art anomaly detection~\cite{laszka2019detection}.

\subsubsection{Attack Baselines}

In \textbf{Greedy} \colorlegendbox{attack} from Eghtesad \emph{et al.}~\shortcite{eghtesad2024multi}, the adversarial agent counts the number of vehicles $s_e^t$ passing through each edge $e$ as their unperturbed shortest path to their destination at the current time step $t$. Then, it divides the budget $B$ proportionally to the number of vehicles:
$\vect{a}^t = \frac{\langle s_e^t :\;\forall e \in E\rangle}{\sum_{e \in E} s_e^t} \cdot B.$

In \textbf{Gaussian} \colorlegendbox{attack} from Yu \emph{et al.}~\shortcite{yu2024sensing}, the attacker divides the environment into multiple subcomponents; we use $k$-means graph clustering \cite{eghtesad2024multi} to divide the environment into subcomponents. The attacker then applies a normal Gaussian perturbation to a subcomponent $i$, dissuading vehicles from passing through $i$:
\begin{align}
    a^t_e = \begin{cases}
        0 \quad &\text{if} \quad e \not\in i \\
        \sim \mathcal{N}(\hat{B} \cdot c_e, \frac{1}{10}\cdot c_e) \quad &\text{otherwise}
    \end{cases}: \forall_{e \in E}, \nonumber
\end{align}
where $\hat{B}$ is the budget applied to each edge in subcomponent~$i$.

\subsubsection{Defense Baselines}

We adopt an anomaly detection algorithm based on the \textbf{Bayesian} \colorlegendbox{defense} process from Laszka \emph{et al.}~\cite{laszka2019detection}. We collect nominal (i.e., without attack) baseline experiences~$x$. The observed travel time of each edge $\hat{w}_e^t$ becomes a random variable for $t \in |H|$, where $|H|$ is the length of the observation history for the defender. The mean and covariance are calculated over the trajectories to form a multivariate normal distribution:
\begin{align}
    F_{\hat{w}} \sim \begin{cases}
        \text{Mean}(\hat{w}^t_e) = \mathbb{E}_x[\hat{w}^t_e] \\
        \begin{aligned}
            \text{Cov}(\hat{w}_{e^i}^{t_i}, \hat{w}_{e^j}^{t_j}) = \mathbb{E}_x[&(\hat{w}_{e^i}^{t_i}-\mathbb{E}_x[\hat{w}_{e^i}^{t_i}]) \\
            \cdot&(\hat{w}_{e^j}^{t_j}-\mathbb{E}_x[\hat{w}_{e^j}^{t_j}])]
        \end{aligned}
    \end{cases}
\end{align}
The detection decision is based on comparing the likelihood of a history of observations $\vect{\hat{w}_{H}}$ according to this distribution with a threshold likelihood $\tau$: $a^t_d = 0$ if $F_{\hat{w}}(\vect{\hat{w}}^t_{H}) > \tau$, otherwise $a^t_d = 1$.

\subsection{Numerical Results}
\label{sec:analysis}

To demonstrate the effectiveness of our approach, we need to show that the calculated equilibrium attack and defense policies satisfy \cref{eq:equilibrium}. In other words, 1) when the equilibrium attacker is playing against an alternative \textcolor{defense}{defense} \protect\colorlegendbox{defense} policy, it achieves a higher total travel time, and 2) when the equilibrium defender is playing against an alternative \textcolor{attack}{attack} \protect\colorlegendbox{attack} policy, it achieves a lower travel time.

Figure~\ref{fig:comparison} shows the numerical results of our equilibrium strategies against the baseline attacks and defenses. \textcolor{black}{\textit{Nominal}} \protect\colorlegendbox{black} shows the normal total travel time of vehicles without any attacks. Comparison of \textit{Nominal} \protect\colorlegendbox{black} to our approach shows that our equilibrium defender limits total travel time deviations by 35\%, 24\%, and 38\% against the worst-cast \textcolor{attack}{attacker} in the 3x2 GRE, 5x4 GRE, and Sioux Falls, SD networks, respectively. 

Our \textcolor{solution}{equilibrium attack strategy} is 19\%, 11\%, and 22\% (episode samples=64, $p$-value=0.0002) more effective compared to the best (i.e., highest) alternative \textcolor{attack}{attack} baseline in 3x2 GRE graph, 5x4 GRE graph, and Sioux Falls, SD networks, respectively. Our \textcolor{solution}{equilibrium defense strategy} is 4\%, 34\%, and 14\% (episode samples=64, $p$-value=0.0002) more robust compared to the best (i.e., lowest) alternative \textcolor{defense}{defense} baseline.

\change{In this figure, the }{}\textcolor{defense}{No Defense} \protect\colorlegendbox{defense} scenario represents the situation where the attacker executes its equilibrium policy (i.e., a worst-case attacker), but there is no detection mechanism. A maximum achievable total travel time exists because of the simulation's finite time horizon.

\section{Conclusion}
\label{sec:conclusion}

We address FDI attacks in crowdsourced navigation applications by formulating a strategic zero-sum game solved via policy space response oracles using deep reinforcement learning as best-response oracles. Our approach yields a robust detection mechanism that limits travel time deviations to 34\%, outperforming state-of-the-art baselines by 22\%. This game-theoretic framework ensures resilient urban mobility against adaptive, worst-case adversarial threats.

\section*{Acknowledgment}
This material is based upon work supported by the National Science Foundation (NSF) under Awards No. CNS-1952011, CCF-2403758, and IIS-2214141, Office of Naval Research under Award No. N00014-24-1-2663, Army Research Office under Award No. W911NF-25-1-0059, and by the U.S. Department of Energy (DOE) under Award No. DE-EE0011188.
Any opinions, findings and conclusions or recommendations expressed in this material are those of the author(s) and do not necessarily reflect the views of the NSF, ONR, ARO, and the U.S. DOE.
We thank the anonymous reviewers for their feedback on our work and for their suggestions to improve our manuscript.

\bibliographystyle{IEEEtran}
\bibliography{references_simplified}

\appendix

\subsection{Hardware Configuration}

We performed all experiments, including neural network training, on a workstation with 2 AMD EPYC 7763 CPUs, each with 64 cores, 1TB of RAM, and an NVIDIA RTX A5000 GPU with 24GB of VRAM. The CPU is capable of executing 128 concurrent simulations. Since our models are relatively small, we needed only 64GB RAM for simulation and inference and 2GB VRAM for model training.

\subsection{Software Configuration}

The implementation of our simulation is an extension of the source code of Eghtesad \emph{et al.}~\shortcite{eghtesad2024multi}. All of our source code, including our extended simulation and our computational framework, is available under an open-source license.
We developed the environments for the attack and defense oracles using Python in the standard format of Farama Gymnasium (formerly known as OpenAI Gym). %
For the single-agent DRL algorithm, we used Stable Baselines 3~\cite{stable-baselines3}, which internally uses PyTorch %
as its neural operations framework. In addition, we used \textit{scikit-learn} %
to solve the linear program that finds the equilibrium of the subgame $G^i$~\cite{shoham2008multiagent}.

When training a model, we drew multiple trajectories (or experiences) for each update from simulations running concurrently. We moved these trajectories to the GPU for training. When running the PSRO algorithm, obtaining one experience requires executing the other agent's previously trained policy. We duplicated this policy, loaded it for each instance of the environment into the main RAM, and executed it on the CPU.

Each independent DRL algorithm can be executed for simulation, data collection, and training at 879, 281, and 198 steps per second on average for the 3x2, 5x4, and Sioux Falls networks, respectively. The substantial difference comes from the computational cost of simulating the vehicles' routing decisions and movements.

\subsection{Hyperparameters}

We adopted the hyperparameters for the Proximal Policy Optimization (PPO) algorithm, used for both the attack and defense oracles, from the default configurations of the Stable Baselines 3~\cite{stable-baselines3} library, which themselves are based on the well-established defaults from OpenAI Baselines. %
This decision was informed by the comprehensive large-scale empirical study conducted by Andrychowicz~\emph{et. al.}~\shortcite{andrychowicz2020what}. Their findings provide strong evidence that these default parameters represent a highly competitive and robust baseline, making them a suitable and valid choice for our experiments. We provide a detailed breakdown of these parameters in Table~\ref{tab:hyperparameters}.

\begin{table*}[t]
\centering
\caption{Hyperparameters of the Attack Oracle, Defense Oracle, and the Double Oracle (DO) Algorithm}
\label{tab:hyperparameters}
\resizebox{\textwidth}{!}{%
\begin{tabular}{@{}llll@{}}
\toprule
\textbf{Component} & \textbf{Hyperparameter} & \textbf{Value} & \textbf{Source File} \\
\midrule
\multicolumn{4}{l}{\textbf{Attack Oracle Hyperparameters}} \\
\cmidrule(l){1-4}
\multirow{13}{*}{\textbf{Training}} & Total Training Timesteps & 5,000,000 & \texttt{models/double\_oracle/trainer.py} \\
 & Algorithm & Proximal Policy Optimization (PPO) & \texttt{models/trainer.py} \\
 & Learning Rate & 0.0003 (default) & \texttt{stable\_baselines3/ppo/ppo.py} \\
 & PPO Number of Steps (n\_steps) & 50 & \texttt{models/trainer.py} \\
 & Batch Size & 64 (default) & \texttt{stable\_baselines3/ppo/ppo.py} \\
 & PPO Number of Epochs (n\_epochs) & 10 (default) & \texttt{stable\_baselines3/ppo/ppo.py} \\
 & Discount Factor (Gamma) & 0.99 (default) & \texttt{stable\_baselines3/ppo/ppo.py} \\
 & GAE Lambda & 0.95 (default) & \texttt{stable\_baselines3/ppo/ppo.py} \\
 & Clip Range & 0.2 (default) & \texttt{stable\_baselines3/ppo/ppo.py} \\
 & Entropy Coefficient (ent\_coef) & 0.01 & \texttt{models/trainer.py} \\
 & Value Function Coefficient (vf\_coef) & 0.5 (default) & \texttt{stable\_baselines3/ppo/ppo.py} \\
 & Max Gradient Norm & 0.5 (default) & \texttt{stable\_baselines3/ppo/ppo.py} \\
 & Normalize Advantage & True (default) & \texttt{stable\_baselines3/ppo/ppo.py} \\
\cmidrule(l){2-4}
\multirow{3}{*}{\textbf{Neural Network}} & Policy & MlpPolicy & \texttt{models/trainer.py} \\
 & Network Architecture & \texttt{dict(pi=[64, 64], vf=[64, 64])} (default) & \texttt{stable\_baselines3/common/policies.py} \\
 & Activation Function & Tanh (default) & \texttt{stable\_baselines3/common/policies.py} \\
\midrule
\multicolumn{4}{l}{\textbf{Defense Oracle Hyperparameters}} \\
\cmidrule(l){1-4}
\multirow{14}{*}{\textbf{Training}} & Total Training Timesteps & 2,000,000 & \texttt{models/double\_oracle/trainer.py} \\
 & Algorithm & Proximal Policy Optimization (PPO) & \texttt{models/double\_oracle/trainer.py}\\
 & Learning Rate & 0.0003 (default) & \texttt{stable\_baselines3/ppo/ppo.py} \\
 & PPO Number of Steps (n\_steps) & 50 & \texttt{models/double\_oracle/trainer.py} \\
 & Batch Size & 64 (default) & \texttt{stable\_baselines3/ppo/ppo.py} \\
 & PPO Number of Epochs (n\_epochs) & 10 (default) & \texttt{stable\_baselines3/ppo/ppo.py} \\
 & Discount Factor (Gamma) & 0.99 (default) & \texttt{stable\_baselines3/ppo/ppo.py} \\
 & GAE Lambda & 0.95 (default) & \texttt{stable\_baselines3/ppo/ppo.py} \\
 & Clip Range & 0.2 (default) & \texttt{stable\_baselines3/ppo/ppo.py} \\
 & Entropy Coefficient (ent\_coef) & 0.01 & \texttt{models/double\_oracle/trainer.py} \\
 & Value Function Coefficient (vf\_coef) & 0.5 (default) & \texttt{stable\_baselines3/ppo/ppo.py} \\
 & Max Gradient Norm & 0.5 (default) & \texttt{stable\_baselines3/ppo/ppo.py} \\
 & Normalize Advantage & True (default) & \texttt{stable\_baselines3/ppo/ppo.py} \\
\cmidrule(l){2-4}
\multirow{3}{*}{\textbf{Neural Network}} & Policy & MlpPolicy & \texttt{models/double\_oracle/trainer/trainer.py} \\
 & Network Architecture & \texttt{dict(pi=[64, 64], vf=[64, 64])} (default) & \texttt{stable\_baselines3/common/policies.py} \\
 & Activation Function & Tanh (default) & \texttt{stable\_baselines3/common/policies.py} \\
\midrule
\multicolumn{4}{l}{\textbf{Double Oracle (DO) Parameters}} \\
\cmidrule(l){1-4}
\multirow{6}{*}{\textbf{DO Parameters}} & DO Iterations & 10 & \texttt{models/double\_oracle/trainer.py} \\
 & Environment Horizon & 50 & \texttt{models/double\_oracle/trainer.py} \\
 & Number of Parallel Environments (n\_envs) & 128 & \texttt{models/double\_oracle/trainer.py} \\
 & Evaluation Episodes & 50 & \texttt{models/trainer.py} \\
 & Post-Training Testing (Numerical Report) Epochs & 64 & \texttt{models/double\_oracle/trainer.py} \\
 \midrule
\multicolumn{4}{l}{\textbf{Environment Hyperparameters}} \\
\cmidrule(l){1-4}
& Softmin Policy ($\theta$) & 1.0 & \texttt{transport\_env/MultiAgentEnv.py} \\
& K-Means Graph Clustering (n\_components) & 4 & \texttt{models/double\_oracle/trainer.py}\\
& False Positive Cost ($C_f$) & 1.0 & \texttt{transport\_env/AdvEnv.py} \\
& History Size & 5 & \texttt{models/double\_oracle/trainer.py} \\
& GRE generation parameter ($p$, $q$) & 0.6057, 0.3162 & \texttt{generate\_gre\_graph.py} \\
\bottomrule
\end{tabular}%
}
\end{table*}

\subsubsection{Random Seeds}

To manage stochasticity, we adhered to the default random number generators provided by Python, NumPy, %
and PyTorch~\cite{paszke2019pytorch} 
without setting explicit global seeds. This approach ensures that the inherent randomness in the training process, such as weight initialization and environment interactions, is handled by the standard, well-vetted procedures of these libraries. This allows our results to reflect the general performance of the methodology rather than being tied to a specific, potentially fortunate, random seed.

\subsection{Statistical Tests}

To validate the significance of our experimental results, we applied permutation tests to compare the performance of our equilibrium policies with each baseline approach. For each comparison, we collected 64 episodic rewards from both our trained policy and the baseline approach. The permutation test then assessed the null hypothesis that the two sets of rewards were drawn from the same random distribution, i.e., that the difference between our proposed approach and the baseline approach is due to chance only.
For this assessment, we had to calculate a $p$-value, which is the probability of observing a difference as large as the one in our experimental results if the null hypothesis were true. We utilized the robust implementation of this statistical test provided by the SciPy %
library. This non-parametric approach is particularly well-suited for our analysis since it does not rely on any assumptions about the underlying distribution of the utilities. %
For every comparison, our test rejected the null hypothesis, demonstrating that the difference between our proposed approach and the baseline approach is statistically significant.

\end{document}